\documentclass{article}

\usepackage{arxiv}

\usepackage[utf8]{inputenc} 
\usepackage[T1]{fontenc}    
\usepackage{hyperref}       
\usepackage{url}            

\usepackage[caption=false]{subfig}

\usepackage[square,numbers]{natbib}
\usepackage[english]{babel}
\usepackage{blindtext}

\usepackage{graphicx}
\usepackage{url}
\usepackage{amsmath,amssymb,mathtools}
\usepackage{multirow}

\DeclareMathOperator*{\argmin}{arg\,min}

\newcommand{\wbest}{h_{\text{best}}}

\usepackage{booktabs}

\newcommand{\hide}[1]{}

\newcommand{\MTSimple}{$\text{M}_{\text{SIMPLE}}$}
\newcommand{\MT}{$\text{M}_{\text{HT}}$}
\newcommand{\MTCV}{$\text{M}_{\text{CV}}$}

\usepackage[ruled,linesnumbered]{algorithm2e}

\title{Making Learners (More) Monotone}

\author{
  Tom Viering\\
  Delft University of Technology, The Netherlands\\
  \texttt{t.j.viering@tudelft.nl} \\
   \And
  Alexander Mey \\
  Delft University of Technology, The Netherlands\\
  \texttt{a.mey@tudelft.nl} \\
  \AND
  Marco Loog \\
  Delft University of Technology, The Netherlands\\
  University of Copenhagen, Denmark\\
  \texttt{m.loog@tudelft.nl} \\
}

\newenvironment{proof}{\par\noindent{\it Proof\ }}{\hfill$\square$}

\newtheorem{theorem}{Theorem}

\newtheorem{definition}{Definition}

\begin{document}
\maketitle

\begin{abstract}
Learning performance can show non-monotonic behavior. That is, more data does not necessarily lead to better models, even on average. We propose three algorithms that take a supervised learning model and make it perform more monotone. We prove consistency and monotonicity with high probability, and evaluate the algorithms on scenarios where non-monotone behaviour occurs. Our proposed algorithm $\text{MT}_{\text{HT}}$ makes less than $1\%$ non-monotone decisions on MNIST while staying competitive in terms of error rate compared to several baselines. 
\end{abstract}

\section{Introduction}

It is a widely held belief that more training data usually results in better generalizing machine learning models --- see for example popular machine learning textbooks on the topic \citep{shalev2014understanding}. Several learning problems have illustrated that more training data can lead to worse generalization performance \citep{Duin1995,Opper1996,Loog2012}. For the peaking phenomema \citep{Duin1995}, this occurs exactly at the transition from the underparametrized to the overparametrized regime. This phenomena has regained interest in the machine learning community in the context of deep neural networks \citep{Belkin2019,Spigler2018}, since these models are typically overparametrized. Recently, also several new examples have been found, where in quite simple settings more data results in worse generalization performance \citep{Viering2019,Loog2019}. 

In practice, it would be very tough to explain why a machine learning model would perform worse when more, typically expensive to collect, data has been used for training. Besides that point,  it seems generally desireable to have algorithms that guarantee increased performance with more data. How to get such a guarantee? This is the question we investigate in this work. 

This is studied using a learning curve: a curve that plots the expected performance of a learning algorithm versus the amount of training data. \footnote{Not to be confused with training curves from deep learning, where the loss versus epochs (optimization iteration) is plotted.} In that context the question becomes: how can we make the learning curve monotonic? 

The core requirement to make learners monotone is that, when more data is gathered and a new model is trained, this newly trained model should be compared to the older model that was trained on less data. And only if the new model performs better should it be used.  We introduce several wrapper algorithms for supervised classification techniques that use the holdout set or cross-validation to do this comparison. Our proposed algorithm $\text{MT}_{\text{HS}}$ uses a hypothesis test to switch if the new model improves significantly upon the old model. Using guarantees from the hypothesis test we can prove that the resulting learning curve is monotone with high probability. We empirically study the effect of the parameters of the algorithms and benchmark them on several datasets including MNIST \citep{Lecun1998} to check to what degree the learning curves become monotone.

This paper is organized as follows. The setting and the concept of monotonicity of learning curves is reviewed in Section \ref{sec_monotone}. We introduce the algorithms in Section \ref{sec_alg}, and prove consistency and monotonicity with high probability in Section \ref{sec_theory}. Section \ref{sec_results} provides the empirical evaluation. We discuss the main findings of our results in Section \ref{sec_discussion} and end with the most important conclusions. 

\section{The Setting and the Definition of Monotonicity} \label{sec_monotone}

We consider the setting where we have a learner that now and then receives data and that is evaluated over time. The question is then, how to make sure that the performance of this learner over time is monotone---or with other words, how can we guarantee that this learner over time improves its performance?   

We analyze this question in a (frequentist) classification framework. We assume there exists an (unknown) distribution $P$ over $\mathcal{X} \times \mathcal{Y}$, where $\mathcal{X}$ is the input space (features) and $\mathcal{Y}$ is the output space (classification labels). To simplify the setup we operate in rounds indicated by $i$, where $i \in \{1,\ldots,n\}$. In each round, we receive a batch of samples $S^i$ that is sampled i.i.d. from $P$. The learner can use this data in combination with data from previous rounds to come up with a hypothesis $h_i$ in round $i$. The hypothesis comes from a hypothesis space $\mathcal{H}$. We consider learners that, as subroutine, use a supervised learner $A: \mathcal{S} \rightarrow \mathcal{H}$, where $\mathcal{S}$ is the space of all possible training sets.

We measure performance by the error rate. The true error rate on $P$ equals 
$$\epsilon(h_i) = \int_{x \in \mathcal{X}} \sum_{y \in \mathcal{Y}} l_{0-1} (h_i(x), y) dP(x,y)$$
where $l_{0-1}$ is the zero-one loss. We indicate the empirical error rate of $h_i$ on a sample $S$ as $\hat{\epsilon}(h_i,S)$. We call $n$ rounds a run. All the ${\epsilon}_i$'s of a run form a learning curve. Averaging multiple runs gives the expected learning curve, $\bar{\epsilon}_i$.

The goal for the learner is twofold. The error rates of the returned models ${\epsilon}_i$'s should (1) be as small as possible, and (2) be monotonically decreasing. 
These goals may be at odds with another; for example, always returning a fixed model ensures monotonicity but incurs large error rates.
To measure (1), we summarize performance of a learning curve using the Area Under the Learning Curve (AULC) \citep{o2017model,huijser2017active,settles2008analysis}. The AULC averages all $\epsilon_i$'s of a run. Low AULC indicates that a learner manages to quickly reduce the error rate.

Monotone in round $i$ means that ${\epsilon}_{i+1} \leq {\epsilon}_i$. 
We may care about monotonicity of the expected learning curve \textit{or} individual learning curves. In practice, we may only get one chance to gather data and submit models. In that case, we would rather want to make sure that then any additional data also leads to better performance during a single run. Therefore, we are mainly concerned with monotonicity of \textit{individual} learning curves. We quantify monotonicity of a run by the fraction of non-monotone transitions in an individual curve.

\section{Algorithms} \label{sec_alg}

We will introduce three algorithms that wrap around supervised learners with the aim of making them monotone. First, we will provide some intuition how to achieve this: ideally, during the generation of the learning curve, we would check whether $\epsilon(h_{i+1}) \leq \epsilon(h_i)$. A fix to make a learner monotone would be to output $h_{i}$ instead of $h_{i+1}$ if the error rate of $h_{i+1}$ is larger. Since learners do not have access to $\epsilon(h_i)$, we have to estimate it using the incoming data. The first two algorithms, $\text{MT}_{\text{SIMPLE}}$ and $\text{MT}_{\text{HT}}$, use the holdout method to this end; newly arriving data is partitioned into training and validation sets. The third algorithm, $\text{MT}_{\text{CV}}$, makes use of cross validation.

\subsection{$\text{MT}_{\text{SIMPLE}}$: Monotone Simple} 

\begin{figure}[htb]
\begin{algorithm}[H]
\SetAlgoLined
\DontPrintSemicolon
\SetKwInOut{Input}{input}
\Input{supervised learner $A$, rounds $n$, batches $S^i$\\$u \in \{$updateSimple, updateHT$\}$\\if $u=$updateHT: confidence level $\alpha$, hypothesis test $HT$}
\BlankLine
\SetAlgoLined
$S_t = \{\}$\;
\For{$i = 1,\ldots,n$}{
Split $S^i$ in $S_t^i$ and $S_v^i$\;
Append to $S_t: S_t = [S_t;S_t^i]$\;
$h_i \leftarrow A(S_t)$\;
$Update_i \leftarrow$ u($S_v^i$, $h_i$, $\wbest$, $\alpha$, $HT$)\;
Append to $S_t: S_t = [S_t;S_v^i]$\; \label{line_append_v_to_t}
\If{$Update_i$ or $i = 1$}{
  $\wbest \leftarrow h_i$\;
}
Return $\wbest$ in round $i$
}
\caption{\MTSimple~ and \MT} \label{alg_Combined}
\end{algorithm}

\noindent\makebox[\textwidth][c]{
\begin{minipage}[t]{0.42\textwidth}
\begin{function}[H]
\SetAlgoLined
\DontPrintSemicolon
\SetKwInOut{Input}{input}
\Input{$S_v^i$, $h_i$, $\wbest$}
\BlankLine
\SetAlgoLined
$P_{current} \leftarrow \hat{\epsilon}(h_i,S_v^i)$\;
$P_{best} \leftarrow \hat{\epsilon}(\wbest,S_v^i)$\; \label{line_update_Pbest}
return $(P_{current} \leq P_{best})$\;
\caption{UpdateSimple()} \label{fcn_S}
\end{function}
\end{minipage}
\hfill
\begin{minipage}[t]{0.57\textwidth}
\begin{function}[H]
\SetAlgoLined
\DontPrintSemicolon
\SetKwInOut{Input}{input}
\Input{$S_v^i$, $h_i$, $\wbest$, confidence level $\alpha$, hypothesis test $HT$}
\BlankLine
\SetAlgoLined
$p = HT(S_v^i,h_i,\wbest) \tcp*[f]{p-value}$\;
return $(p \leq alpha)$\;
\caption{UpdateHT()} \label{fcn_HT}
\end{function}
\end{minipage}
}
\caption{The algorithm combined with UpdateSimple gives $\text{MT}_{\text{SIMPLE}}$, the algorithm combined with UpdateHT gives $\text{MT}_{\text{HT}}$. Note that $\text{MT}_{\text{HT}}$ requires additional input parameters $\alpha$ and $HT$, which are not needed by $\text{MT}_{\text{SIMPLE}}$.}
\end{figure}

The pseudo-code for $\text{MT}_{\text{SIMPLE}}$ is given by Algorithm \ref{alg_Combined} in combination with the update function \ref{fcn_S}. Batches $S^i$ are split into training ($S_t^i$) and validation ($S_v^i$). The training set $S_t$ is enlarged each round with $S_t^i$ and a new model $h_i$ is trained. $S_v^i$ is used to estimate the performance of $h_i$. At all times the algorithm stores the previously best performing model, $\wbest$, and compares its performance to that of $h_i$. If the new model $h_i$ is better, it is returned in round $i$ and $\wbest$ is updated, otherwise the algorithm returns $\wbest$.  

In each iteration the performance estimate of $\wbest$ is also updated (see line \ref{line_update_Pbest} in  \ref{fcn_S}) using $S_v^i$. Thus $h_i$ and $\wbest$ are both compared on $S_v^i$, resulting in a more accurate comparison (because the comparison is paired). After the comparison $S_v^i$ can safely be added to the training set (line \ref{line_append_v_to_t} of Algorithm \ref{alg_Combined}). 

We call this algorithm $\text{MT}_{\text{SIMPLE}}$ because the model selection is a bit naive: for small validation sets, the variance in the performance measure could be quite large, leading to many non-monotone decisions. In the limit of infinitely large $S_v^i$, however, this algorithm should always be monotone (and very data hungry).

\subsection{$\text{MT}_{\text{HT}}$: Monotone Hypothesis Test} 

The second algorithm, $\text{MT}_{\text{HT}}$, aims to resolve the issues of $\text{MT}_{\text{SIMPLE}}$ with small validation set sizes. In addition, for this algorithm, we will later prove that individual learning curves are monotone with high probability. The same pseudo-code is used as for $\text{MT}_{\text{SIMPLE}}$ (Alorithm \ref{alg_Combined}), but with a different update function \ref{fcn_HT}. Now a hypothesis test $HT$ is determines if the newly trained model is significantly better than the previous model. The hypothesis test makes sure that the newly trained model is not better due to chance (such as an unlucky sample). 
The hypothesis test is conservative, and only switches to a new model if we are reasonably sure it is significantly better, to avoid non-monotone decisions. \citet[chap. 2.2.3-2.2.4]{japkowicz2011evaluating} provides an accessible introduction to understand the frequentist hypothesis testing framework for machine learners. 

The choice of hypothesis test depends on the performance measure. For the error rate McNemar's test can be used (see experimental setup for more details) \citep{japkowicz2011evaluating,raschka2018model}. For the hypothesis test, there are several requirements: it should use paired data, since we evaluate two models on one sample, and it should be one-tailed.
One-tailed, since we only want to know whether $h_i$ is better than $\wbest$ (a two tailed test would switch to $h_i$ if its performance is significantly different, which is not what we want). Thus we have two hypotheses: $H_0: \epsilon(h_i) = \epsilon(\wbest)$ and $H_1: \epsilon(h_i) \leq \epsilon(\wbest)$. 

We judge significance using the $p$-value: the probability of observing a more extreme sample given hypothesis $H_0$. The smaller the $p$-value, the more evidence we have for $H_1$. The confidence level $\alpha \in (0,\frac{1}{2}]$ indicates the threshold. If the $p$-value is smaller than $\alpha$, we accept $H_1$, and thus we update the model $\wbest$. The smaller $\alpha$, the more conservative the hypothesis test, and thus the smaller the chance that a wrong decision is made due to unlucky sampling. More precisely, most hypothesis tests satisfy that the False Positive Rate (FPR, or the probability to make a Type I error) is bounded: $P(p \leq \alpha | H_0) \leq \alpha$. 
\subsection{$\text{MT}_{\text{CV}}$: Monotone Cross Validation}
In practice, often $K$-fold cross validation (CV) is used to estimate model performance instead of the holdout. This is what $\text{MT}_{\text{CV}}$ does, and is similar to $\text{MT}_{\text{SIMPLE}}$. As described in Algorithm \ref{alg_MTCV}, for each incoming sample an index $I$ maintains to which fold it belongs. These indices are used to generate the folds for the $K$-fold cross validation. 

During CV, $K$ models are trained and evaluated on the validation sets. We now have to memorize $K$ previously best models, one for each fold. We average the performance of the newly trained models over the $K$-folds, and compare that to the average of the best previous $K$ models. This averaging over folds is essential, as this reduces the variance of the model selection step as compared to $\text{MT}_{\text{SIMPLE}}$. As with $\text{MT}_{\text{SIMPLE}}$ paired samples are used for the comparison.  

After the comparison we know which training size was better. Our framework requires us to return a single model in each iteration. We choose to return the model with the optimal training set size that performed best during CV, as this may further improve the performance.

\begin{algorithm}[H]
\SetAlgoLined
\DontPrintSemicolon
\SetKwInOut{Input}{input}
\Input{$K$ folds, learner $A$, rounds $n$, batches $S^i$}
\BlankLine
\SetAlgoLined
$b \leftarrow 1$ \tcp*[f]{keeps track of best round}\; 
$S = \{\}$, $I = \{\}$\;
\For{$i = 1,\ldots,n$}{
Generate stratified CV indices for $S^i$ and put in $I^i$. Each index in indicates to which validation fold the corresponding sample belongs.\;
Append to $S$: $S \leftarrow [S; S^i]$\;
Append to $I$: $I \leftarrow [I; I^i]$\;
\For{$k = 1,\ldots,K$}{
$h_i^k \leftarrow A(S[(I \neq k)])$\tcp*[f]{training set of $k$th fold}\;  
$P_{i}^k \leftarrow \hat{\epsilon}(h_i^k,S[I = k])$ \tcp*[f]{validation set of $k$th fold}\;  
$P_{b}^k \leftarrow \hat{\epsilon}(h_b^k,S[I = k])$\tcp*[f]{update performance of prev. models}\; \label{line_update_Pbest_again}
}
$Update_i \leftarrow (mean(P_{i}^k) < mean(P_b^k))$ \tcp*[f]{mean w.r.t. $k$}\; 
\If{$Update_i$ or $i=1$}{
$b \leftarrow i$\;
}
$k \leftarrow \argmin_{k} P_{b}^k$\tcp*[f]{break ties} \; 
Return $h_{b}^{k}$ in round i\;
}
\caption{\MTCV} \label{alg_MTCV}
\end{algorithm}

\section{Theoretical Analysis} \label{sec_theory}

In this section we derive the probability of a monotone learning curve for the algorithms $\text{MT}_{\text{SIMPLE}}$ and $\text{MT}_{\text{HT}}$, and we prove that all algorithms are consistent as long as they are guaranteed to update the model enough. 
\begin{theorem} \label{thm_highprobability}
Assume that the hypothesis test $HT$ satisfies $P(p \leq \alpha | H_0) \leq \alpha$. Then running Algorithm $\text{MT}_{\text{HT}}$ with parameter $\alpha$ guarantees that the individual learning curve of $n$ rounds is monotone with probability $(1-\alpha)^n$.
\begin{proof}
The probability of making a non-monotone decision in round $i$ is at most $\alpha$, this is guaranteed by the hypothesis test. To see this, assume $\epsilon(h_i) \geq \epsilon(\wbest)$. Let $p$ be the $p$-value as returned by $HT$ as before. The probability of accepting becomes larger if $H_1$ is true $P(p \leq \alpha | H_1) \geq P(p \leq \alpha | H_0)$. From this it should be clear that if $\epsilon(h_i) \geq \epsilon(\wbest)$, the probability of accepting will be even smaller: $P(p \leq \alpha| \epsilon(h_i) \geq \epsilon(\wbest)) \leq P(p \leq \alpha | H_0)$. In a worst case $\epsilon(h_i) \geq \epsilon(\wbest)$ holds every round. Note that these guarantees on the probability of failure hold for any model $h_i$, $\wbest$ and anything that happened before round $i$. Since $S_v^i$ are independent samples, being non-monotone in each round can be seen as independent events, thus we can multiply the probabilities resulting in $(1-\alpha)^n$.
\end{proof}
\end{theorem}
If the probability of being non-monotone in all rounds may be at most $\beta$, we may set $\alpha=1-\beta^{\frac{1}{n}}$ to fulfill this condition. Note that this analysis also holds for $\text{MT}_{\text{SIMPLE}}$, since running $\text{MT}_{\text{HT}}$ with $\alpha=\frac{1}{2}$ results in the same algorithm as $\text{MT}_{\text{SIMPLE}}$ if $HT$ satisfies $P(p \leq \alpha | H_0) \leq \alpha$. 
Now will argue that all proposed algorithms are consistent under some conditions. First we revisit the definition of consistency as defined by \citet{shalev2014understanding}. 
\begin{definition}[Consistency\citep{shalev2014understanding}]
Let $\mathcal{H}$ be the hypothesis class and let $A$ be the learner. For all $\epsilon \in (0,1)$, for all distributions $D$ over $X \times Y$, for all $\delta \in (0,1)$, if there exists a $n(\epsilon,D,\delta)$, such that for all $m \geq n(\epsilon,D,\delta)$, if $A$ is trained on a sample $S$ of size $m$, and the following holds with probability (over the choice of $S$) at least $1-\delta$,
\begin{equation}
    L_D(A(S)) \leq \min_{h \in \mathcal{H}} L_D(h) + \epsilon,
\end{equation}
then $A$ is said to be consistent.
\end{definition}
Before we can state the main result, we have to introduce a bit of notation. $U_i$ will indicate the event that the algorithm updates $\wbest$ (or in case of $\text{M}_{\text{CV}}$ it will update the variable $b$). We will indicate $H_{i}^{i+z}$ to indicate the event that $\neg U_i \cap \neg U_{i+1} \cap \ldots \cap \neg U_{i+z}$, or in words, that in round $i$ to $i+z$ there has been no update. 
To fulfill consistency, we need that when the number of rounds grows to infinity, the probability of updating will be large enough. Then consistency of $A$ will make sure that $\wbest$ has sufficiently low error. For this analysis it is assumed that the number of rounds of the algorithms is not fixed. 
\begin{theorem} \label{thm_consistency}
The algorithms $\text{MT}_{\text{SIMPLE}}$, $\text{MT}_{\text{HT}}$ and $\text{MT}_{\text{CV}}$ are consistent, if $A$ is consistent and if there exists a $C_z>0$ such that for all $i$ we have $P(H_i^{i+z}) \leq (1 - C_z)^z$.
\begin{proof}
Let $A$ be consistent with $n_A(\epsilon,D,\delta)$ samples. Let us analyze round $i$ where $i$ is big enough such that\footnote{In case of $\text{MT}_{\text{CV}}$, take $|S_t|$ to be the smallest training fold size in round $i$} $|S_t| > n_A(\epsilon,D,\frac{\delta}{2})$. Assume that 
\begin{equation}
L(\wbest) > \min_{h \in \mathcal{H}} L(h) + \epsilon,
\end{equation}
otherwise the proof is trivial. 
Since $|S_t| > n_A(\epsilon,D,\frac{\delta}{2})$, we have for any round $j \geq i$ that 
\begin{equation}
L(h_j) \leq \min_{h \in \mathcal{H}} L(h) + \epsilon
\end{equation}
holds with probability of at least $1-\frac{\delta}{2}$. Thus now the algorithm should update. The probability that in the next $z$ rounds we don't update is, by assumption, bounded by $(1-C_z)^z$. Since $C_z>0$, there exists a $z \geq 1$ such that $(1-C_z)^{z} \leq \frac{\delta}{2}$. Thus the probability of not updating after $z$ more rounds is at most $\frac{\delta}{2}$, and we have a probability of $\frac{\delta}{2}$ that the model after updating is not good enough. Applying the union bound, we find that the probability of failure is at most $\delta$ as desired. 
\end{proof}

\end{theorem}

A few remarks about the assumption. It tells us, that an update is more and more likely if we have more consecutive rounds where there has been no update. This holds for example if there are enough rounds where the probability of an update is nonzero. A weaker but also sufficient assumption would be that $\forall_i : \lim_{z \rightarrow \infty} p(H_i^{i+z}) \rightarrow 0$. 

For $\text{MT}_{\text{SIMPLE}}$ and $\text{MT}_{\text{CV}}$ the assumption is always satisfied, because these algorithms look directly at the mean error rate --- and due to fluctuations in the sampling there is always a non-zero probability that $\hat{\epsilon}(h_i) \leq \hat{\epsilon}(\wbest)$. However, for $\text{MT}_{\text{HT}}$ this may not always be satisfied. Especially if the validation batches $N_v$ are small, the hypothesis test may not be able to detect small differences in error -- the test then has zero power. If $N_v$ stays small, even in future rounds the power may stay zero, and then the learner is not consistent.

\section{Experiments} \label{sec_results}

We evaluate $\text{MT}_{\text{SIMPLE}}$ and $\text{MT}_{\text{HT}}$ on artificial datasets to understand the influence of their parameters. Afterward we perform a benchmark where we also include $\text{MT}_{\text{CV}}$ and a baseline that uses validation data to tune the regularization strength. This last experiment is also performed on the MNIST dataset to get an impression of the practicality of the proposed algorithms. First we describe the experimental setup in more detail.

\subsection{Experimental Setup} 

The peaking dataset \citep{Duin1995} and dipping dataset \citep{Loog2012} are artificial datasets that cause non-monotone behaviour. We use stratified sampling to obtain batches $S^i$ for the peaking and dipping dataset, for MNIST we use random sampling. For simplicity all batches have the same size. $N$ indicates batch size, and $N_v$ and $N_t$ indicate the sizes of the validation and training sets. 

As model we use least squares classification \citep{hastie2005elements,rifkin2003regularized}. This is ordinary linear least squares regression on the classification labels $\{-1,+1\}$ with intercept. For MNIST one-versus-all is used to train a multi-class model. In case there are less samples for training than dimensions, the required inverse of the covariance matrix is ill-defined and we resort to the Moore-Penrose Pseudo-Inverse. 
 
 Monotonicity is calculated by the fraction of non-monotone iterations per run. AULC is also calculated per run.
We do 100 runs with different batches and average to reduce variation from the randomness in the batches. Each run uses a newly sampled test set consisting of 10000 samples. The test set is used to estimate the true error rate and is not accessible by any of the algorithms.

We evaluate $\text{M}_{\text{SIMPLE}}$, $\text{M}_{\text{HT}}$ and $\text{M}_{\text{CV}}$ and several baselines. The standard learner just trains on all received data. A second baseline, ${\lambda}_S$, splits the data in train and validation like $\text{M}_{\text{SIMPLE}}$ and uses the validation data to select the optimal $L_2$ regularization parameter $\lambda$ for the least square classifier. Regularization is implemented by adding $\lambda I$ to the estimate of the covariance matrix. 

Several versions of McNemar's test can be used to compare models  \citep{Fagerland2013,japkowicz2011evaluating,raschka2018model}. We use the McNemar's exact conditional test \citep{Fagerland2013}, since for this test all assumptions are satisfied, and as such $P(p \leq \alpha | H_0) \leq \alpha$ is guaranteed. 

In the first experiment we investigate the influence of $N_v$ and $\alpha$ for $\text{MT}_{\text{SIMPLE}}$ and $\text{MT}_{\text{HT}}$ on the decisions. A complicating factor is that if $N_v$ changes, not only decisions change, but also training set sizes because $S_v$ is appended to the training set (see line \ref{line_append_v_to_t} of Algorithm \ref{alg_Combined}). This makes interpretation of the results difficult because decisions are then made in a different context. Therefore, for the first set of experiments, we do not add $S_v$ to the training sets, also not for the standard learner. For this set of experiment We use $N_l = 4$, $n=150$, $d=200$ for the peaking dataset, and we vary $\alpha$ and $N_v$. 

For the benchmark, we set $N_l = 10$, $N_v = 40$, $n=150$ for peaking and dipping, and we set $N_l = 5$, $N_v = 20$, $n = 40$ for MNIST. We fix $\alpha = 0.05$ and use $d = 500$ for the peaking dataset. For MNIST, as preprocessing step we extract 500 random Fourier-features as also done by \citet{Belkin2019}. For $\text{MT}_{\text{CV}}$ we use $K=5$ folds. For $\lambda_{S}$ we try $\lambda \in \{10^{-5}, 10^{-4.5}, \ldots, 10^{4.5}, 10^5\}$ for peaking and dipping, and we try $\lambda \in \{10^{-3}, 10^{-2}, \ldots, 10^3\}$ for MNIST.

\subsection{Results}

\begin{figure}[p]
\centering
\subfloat[Peaking learning curve\label{fig_lc_peaking}]{  \includegraphics[width=0.33\textwidth]{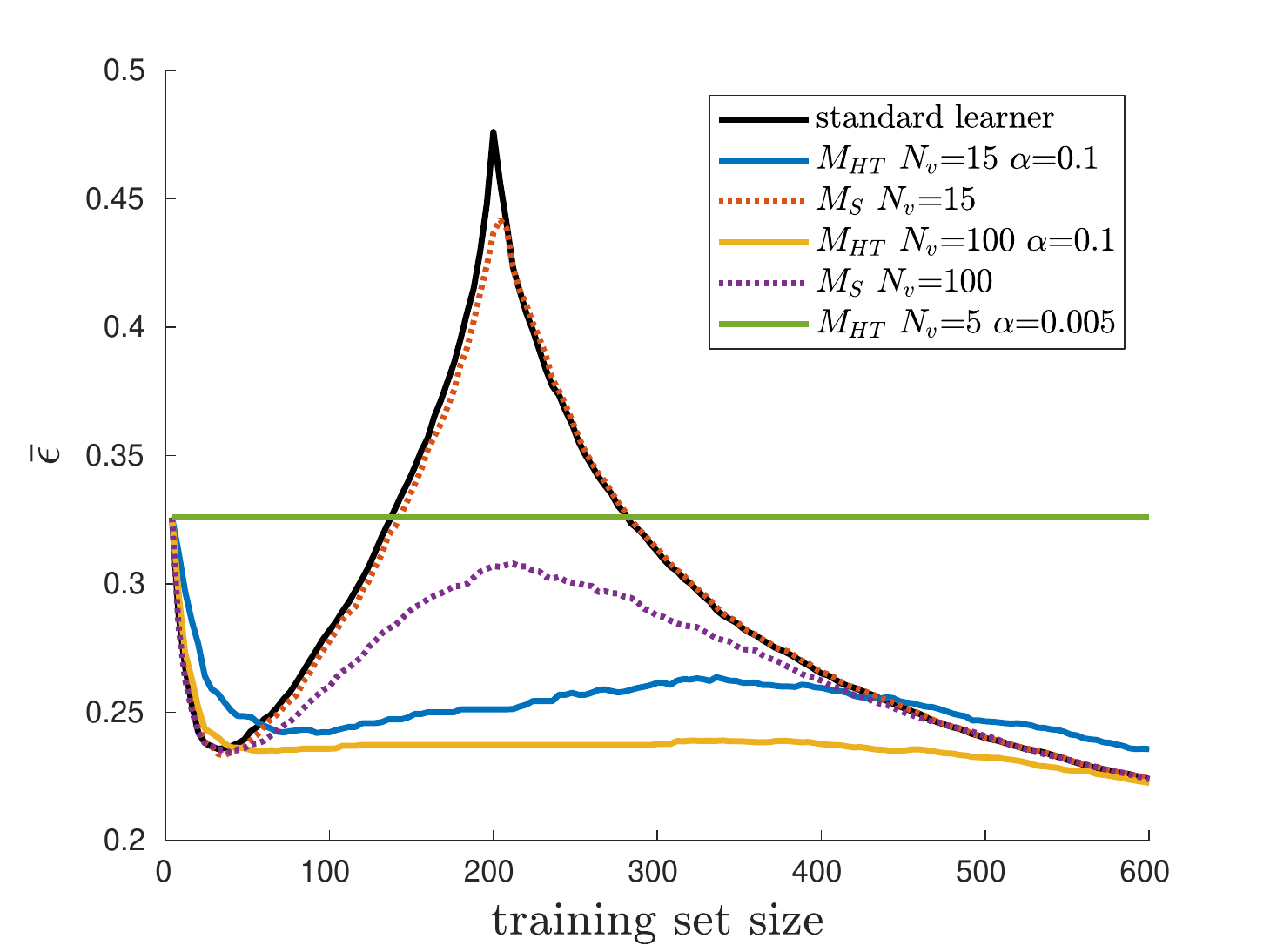}}
\subfloat[Peaking AULC\label{fig_AULC_peaking}]{  \includegraphics[width=0.33\textwidth]{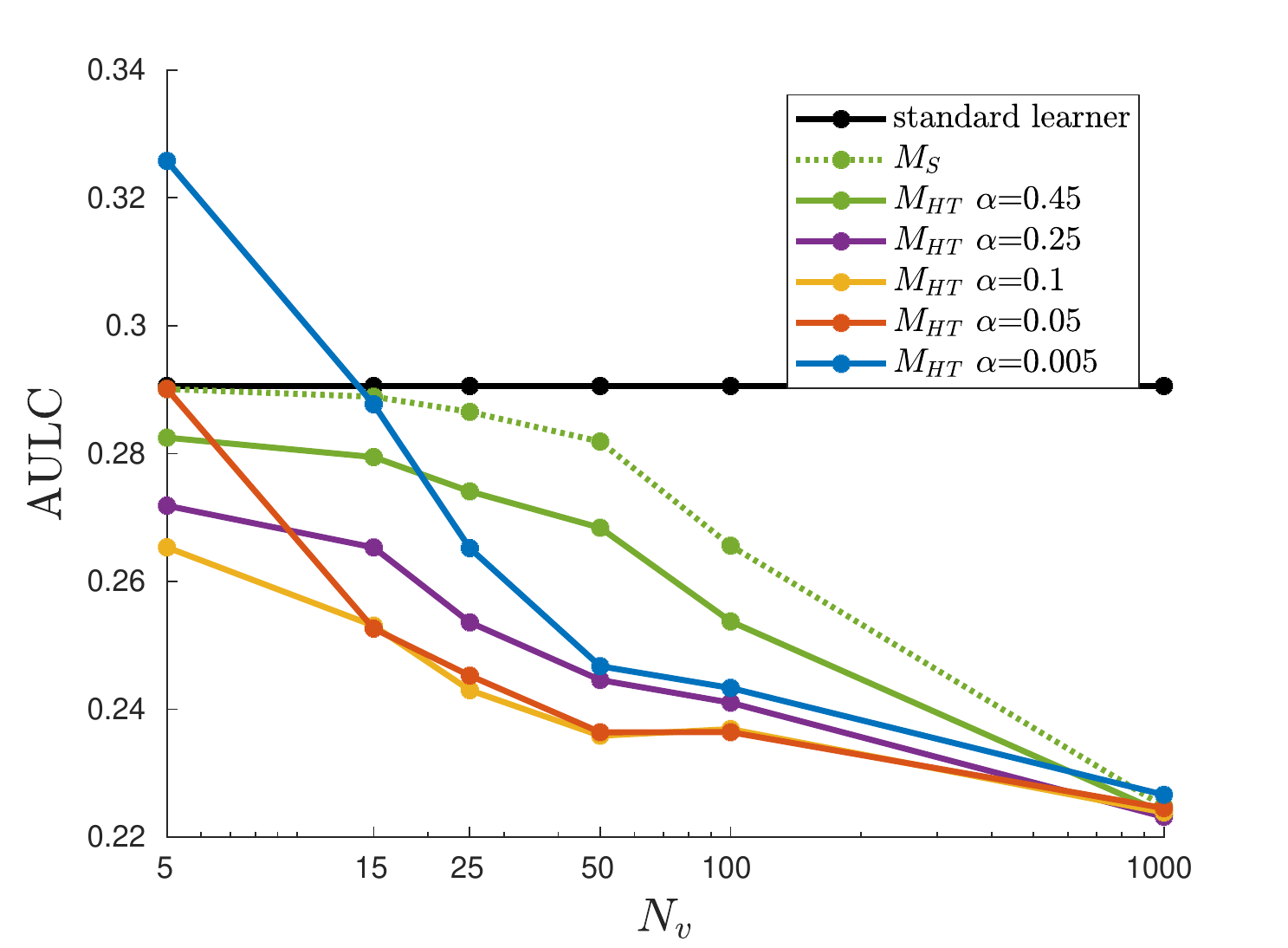}}
\subfloat[Peaking Monotonicity \label{fig_mon_peaking}]{  \includegraphics[width=0.33\textwidth]{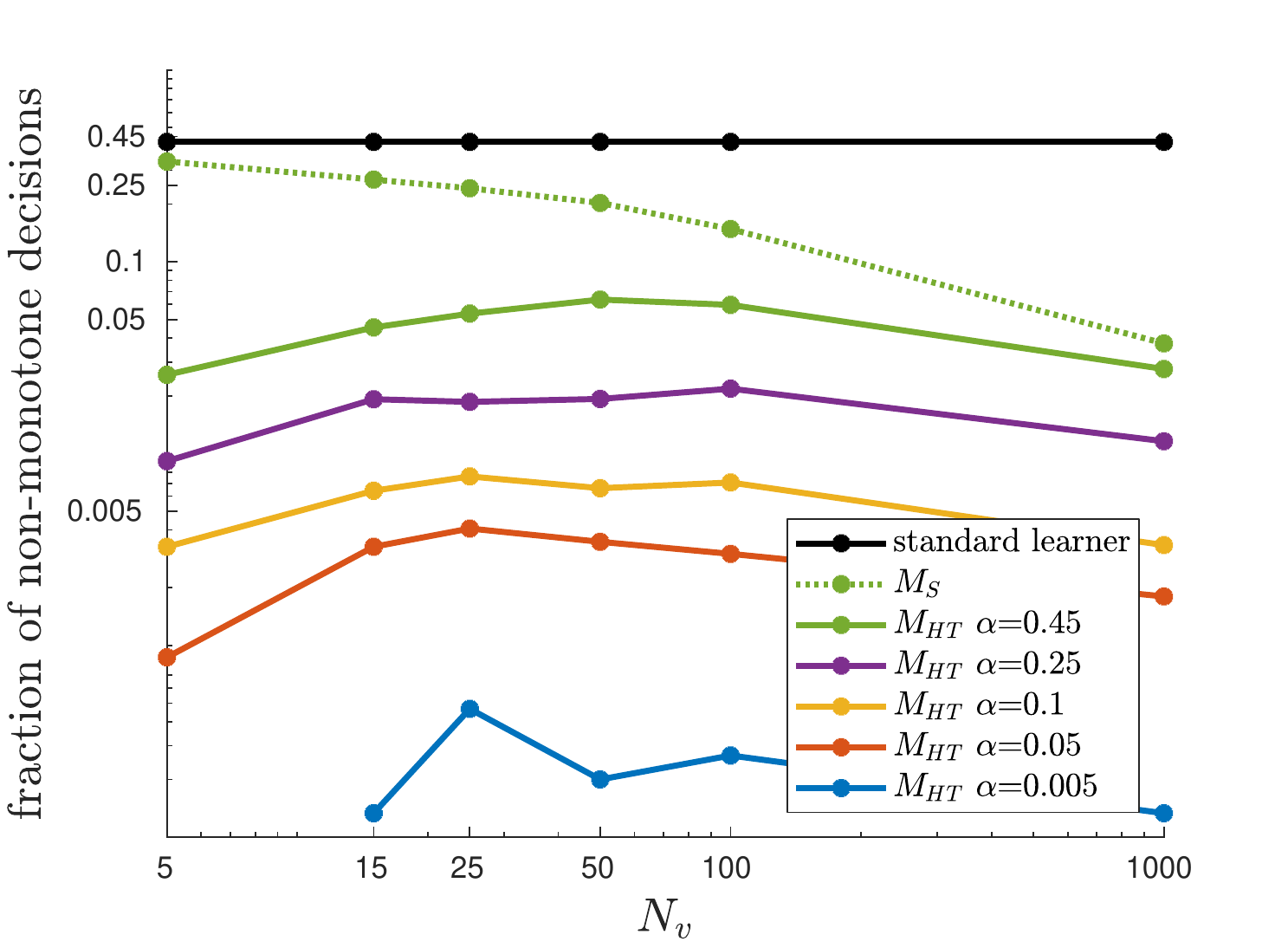}}
\\
\subfloat[Dipping learning curve\label{fig_lc_dipping}]{  \includegraphics[width=0.33\textwidth]{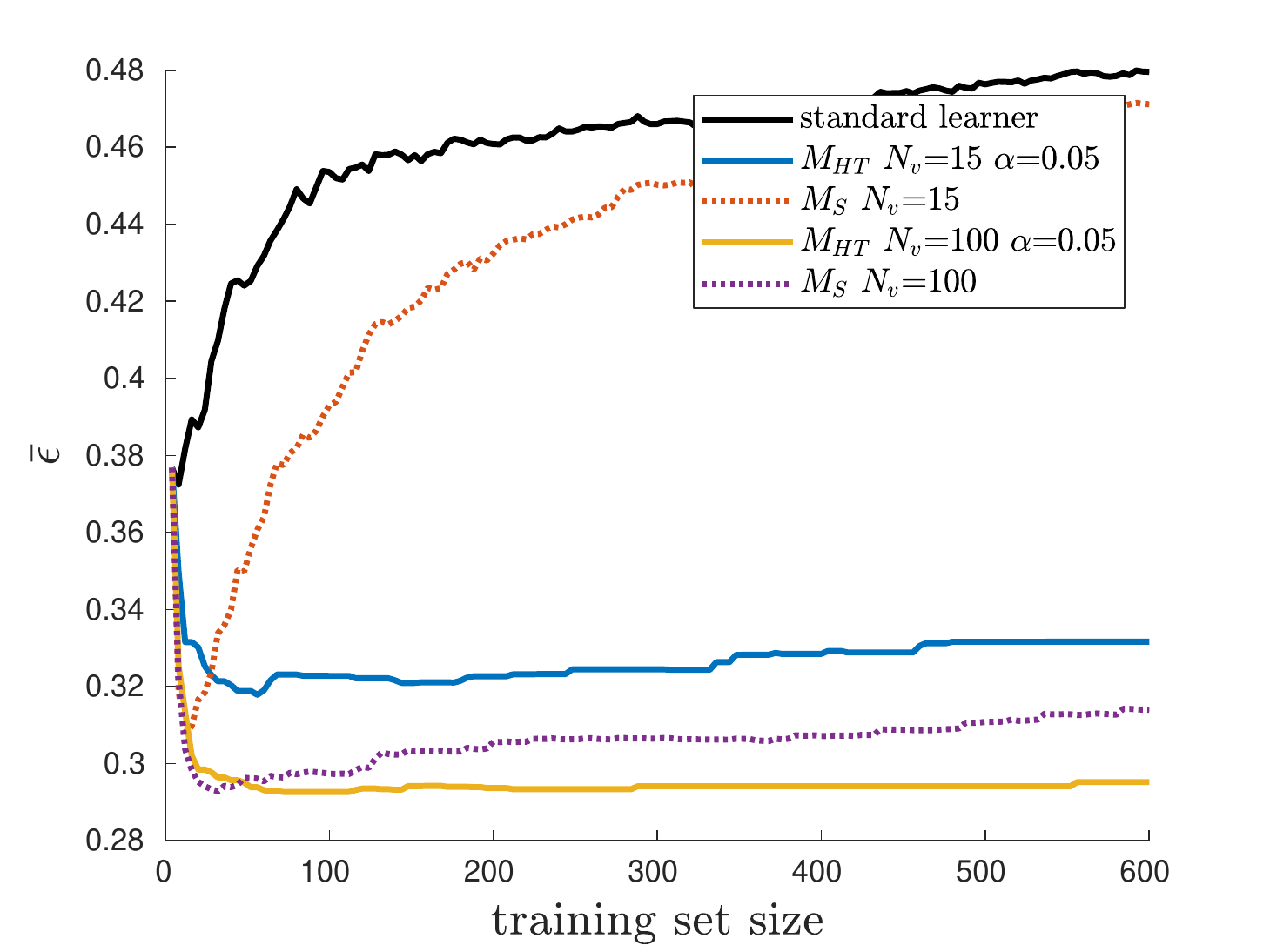}}
\subfloat[Dipping AULC\label{fig_AULC_dipping}]{  \includegraphics[width=0.33\textwidth]{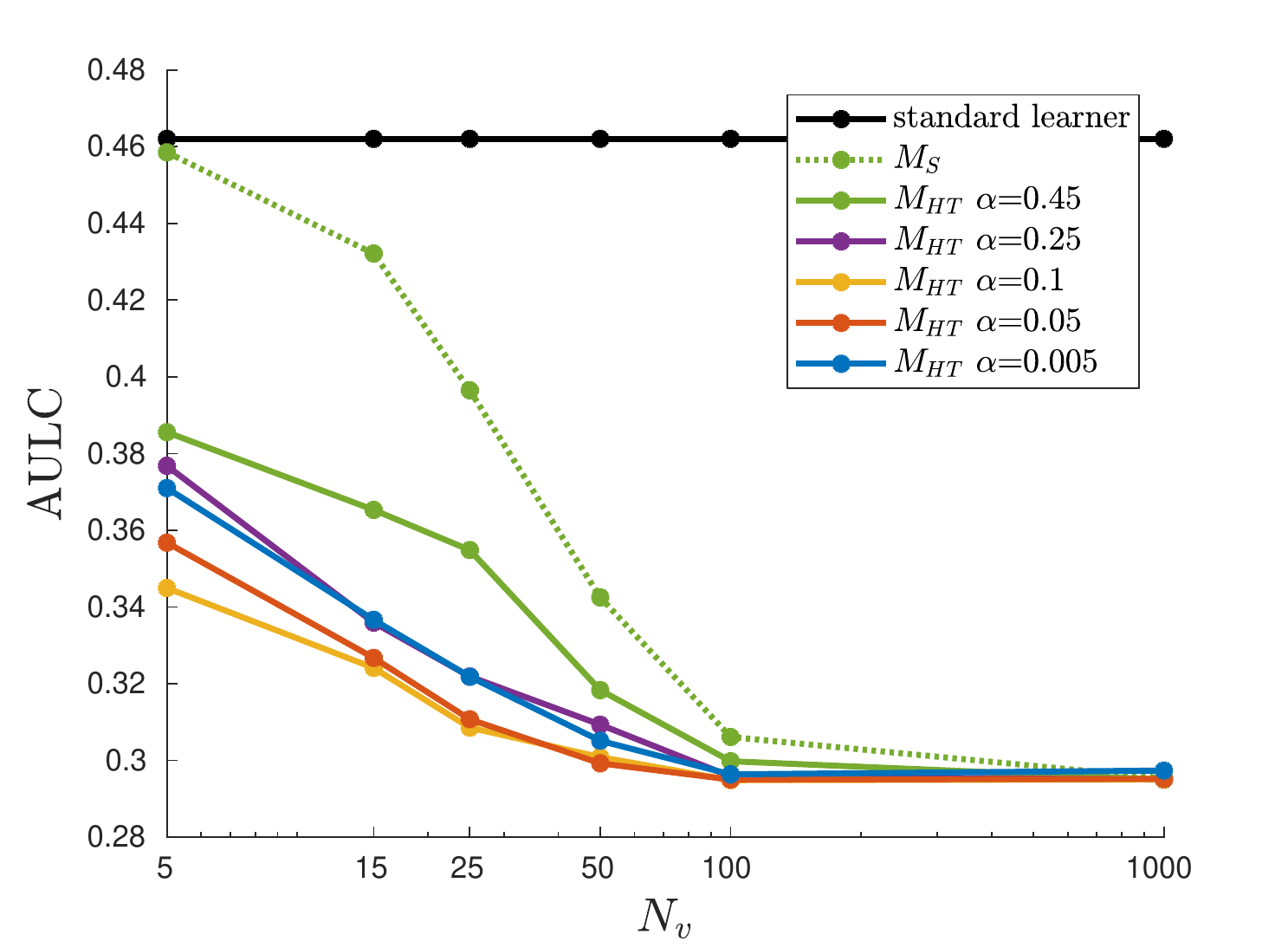}}
\subfloat[Dipping Monotonicity\label{fig_mon_dipping}]{  \includegraphics[width=0.33\textwidth]{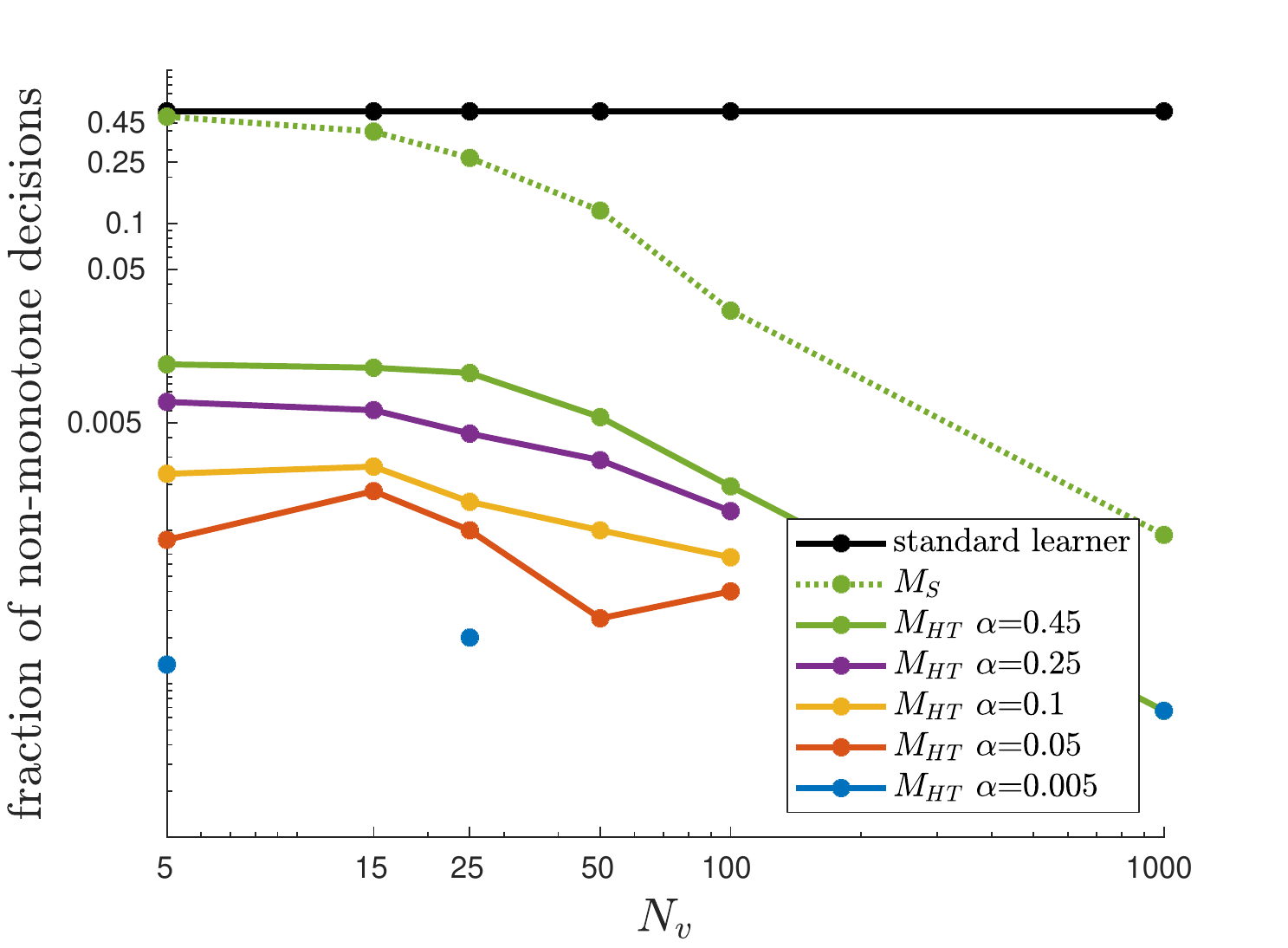}}
\caption{Several experiments on the Peaking and Dipping dataset to investigate the influence of $N_v$ and $\alpha$ for $\text{MT}_{\text{SIMPLE}}$ and $\text{MT}_{\text{HT}}$.}
\label{fig:image2}
\end{figure}

We perform a preliminary investigation of the algorithms $\text{M}_{\text{SIMPLE}}$ and $\text{M}_{\text{HT}}$ and the influence of the parameters $N_v$ and $\alpha$. We show several learning curves in Figure \ref{fig_lc_peaking} and \ref{fig_lc_dipping}. For small $N_v$ and $\alpha$ we observe $\text{MT}_{\text{HT}}$ gets stuck: it does not switch models anymore. This indicates that indeed the assumption required for consistency is not satisfied.

In Figure \ref{fig_AULC_peaking} and Figure \ref{fig_AULC_dipping} we give a more complete picture of all tried hyperparameters in terms of the AULC. In Figure \ref{fig_mon_peaking} and Figure \ref{fig_mon_dipping} we plot the fraction of non-monotone decisions during a run. Observe that this is a log-log plot. In some cases zero non-monotone decisions were observed, and thus the log-log plot misses a value. This occurs for example if $\text{MT}_{\text{HT}}$ always sticks to the same model, then no non-monotone decisions will be made.

\begin{figure}[p]
\centering
\subfloat[Peaking\label{fig_bench_peaking}]{  \includegraphics[width=0.32\textwidth]{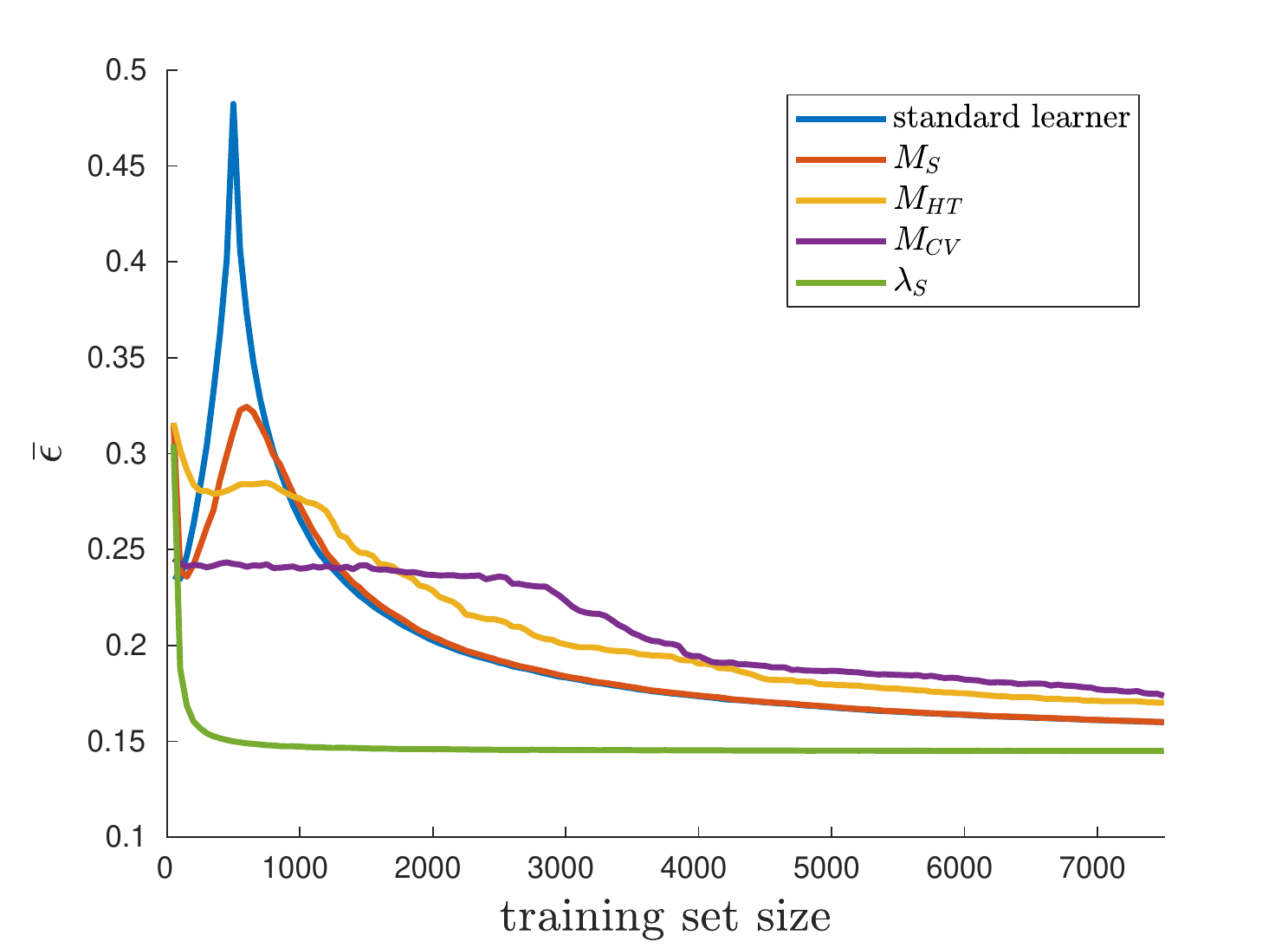}}
\subfloat[Dipping\label{fig_bench_dipping}]{  \includegraphics[width=0.32\textwidth]{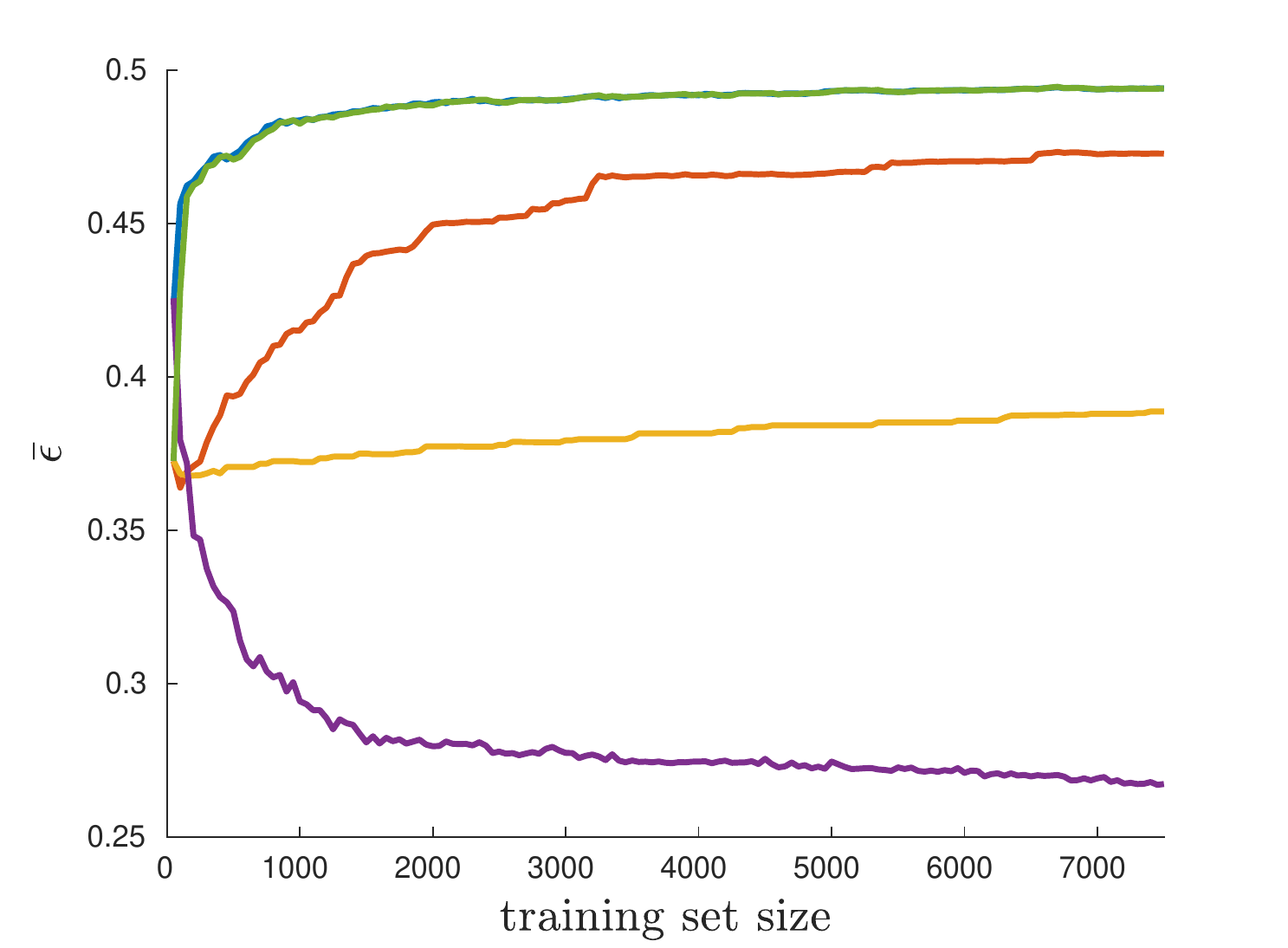}}
\subfloat[MNIST\label{fig_bench_MNIST}]{  \includegraphics[width=0.32\textwidth]{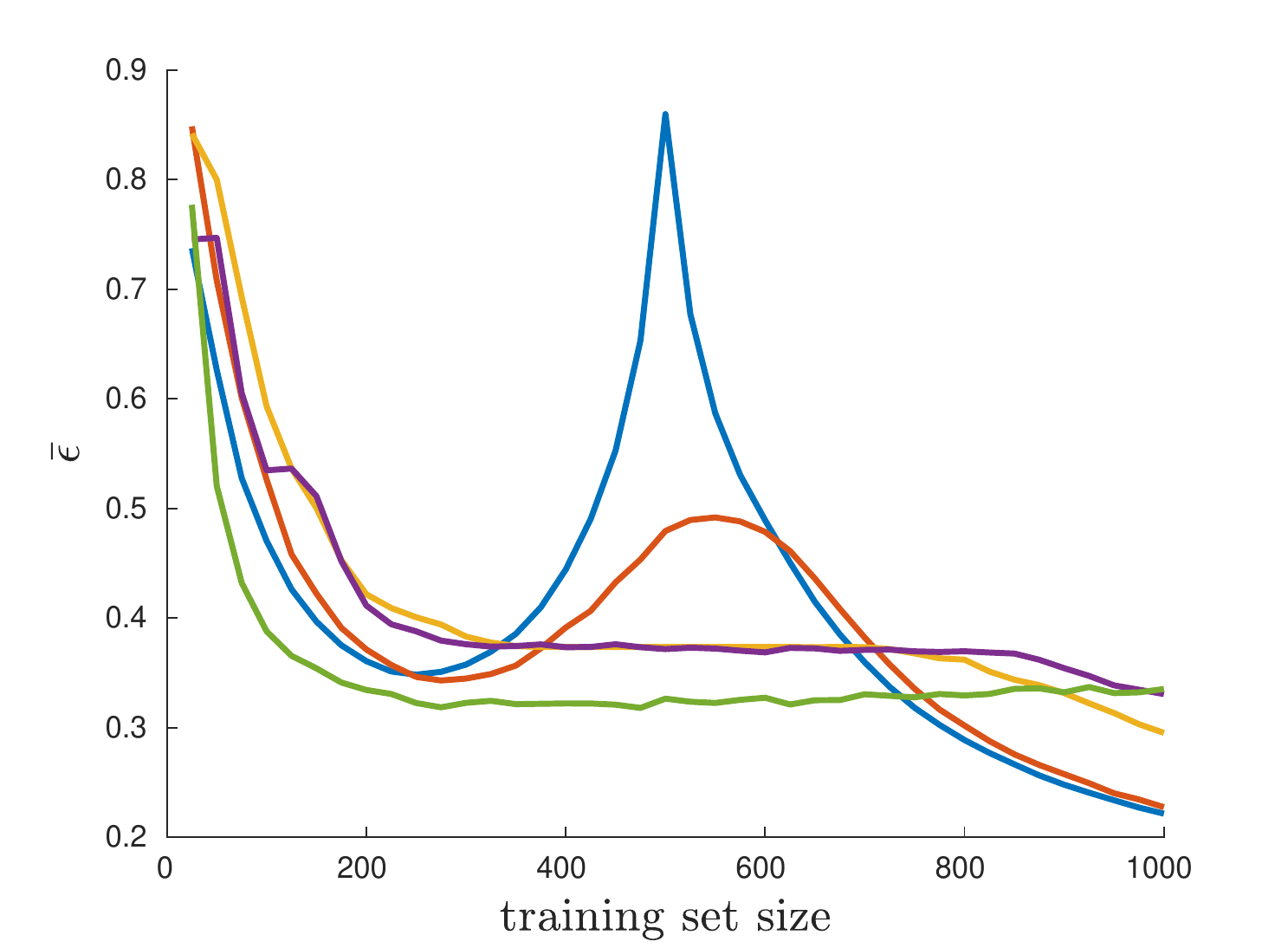}}
\caption{Expected learning curves on the benchmark datasets.}
\label{fig_bench}
\end{figure}

Results of the benchmark are shown in Figure \ref{fig_bench}. The AULC and fraction of monotone decisions are given in Table \ref{table_bench}. 

\section{Discussion} \label{sec_discussion}

\subsection{First experiment: tuning $\alpha$ and $N_v$}

As predicted $\text{MT}_{\text{SIMPLE}}$ typically performs worse than $\text{MT}_{\text{HT}}$ in terms of AULC and monotonicity unless $N_v$ is very large. The variance in the estimate of the error rates on $S_v^i$ is so large that in most cases the algorithm doesn't switch to the correct model. 

Larger $N_v$ leads typically to improved AULC. $\alpha \in [0.05,0.1]$ seems to work best in terms of AULC for most values of $N_v$. If $\alpha$ is too small, $\text{MT}_{\text{HT}}$ can get stuck, if $\alpha$ is too large, it switches models too often and non-monotone behaviour occurs. If $\alpha \rightarrow \frac{1}{2}$, $\text{MT}_{\text{HT}}$ becomes increasingly similar to $\text{MT}_{\text{SIMPLE}}$ as predicted by the theory. 

The fraction of non-monotone decisions of $\text{MT}_{\text{HT}}$ is much lower than $\alpha$. This is in agreement with Theorem \ref{thm_highprobability}, but may indicate that the hypothesis test is rather pessimistic. The standard learner and $\text{MT}_{\text{SIMPLE}}$ often make non-monotone decisions, in some cases almost $50\%$ of the decisions are not-monotone.

\begin{table}[p]
\centering
\footnotesize
\begin{tabular}{lllllll}
\toprule
        & \multicolumn{2}{c}{Peaking} & \multicolumn{2}{c}{Dipping} & \multicolumn{2}{c}{MNIST} \\ \cmidrule(r){2-3}\cmidrule(rl){4-5}\cmidrule(l){6-7}
        & AULC       & Fraction       & AULC       & Fraction       & AULC      & Fraction      \\ \midrule
           SL & 	 \scriptsize\tt 0.198 (0.003) $~$ &	 \scriptsize\tt 0.31 (0.02) $~~$& 	 \scriptsize\tt 0.49 (0.01) $~$ &	 \scriptsize\tt 0.50 (0.03) $~~$& 	 \scriptsize\tt 0.42 (0.01) $~$ &	 \scriptsize\tt 0.27 (0.03) $~~$\\
            $M_{S}$  & 	 \scriptsize\tt 0.195 (0.005) $~$ &	 \scriptsize\tt 0.23 (0.03) $~~$& 	 \scriptsize\tt 0.45 (0.06) $~$ &	 \scriptsize\tt 0.37 (0.15) $~~$& 	 \scriptsize\tt 0.40 (0.02) $~$ &	 \scriptsize\tt 0.12 (0.04) $~~$\\
           $M_{HT}$  & 	 \scriptsize\tt 0.208 (0.009) $~$ &	 \scriptsize\tt \underline{0.00} (0.00) $~~$& 	 \scriptsize\tt 0.38 (0.08) $~$ &	 \scriptsize\tt \underline{0.00} (0.00) $~~$& 	 \scriptsize\tt 0.41 (0.03) $~$ &	 \scriptsize\tt \underline{0.00} (0.01) $~~$\\
           $M_{CV}$  & 	 \scriptsize\tt 0.208 (0.005) $~$ &	 \scriptsize\tt 0.34 (0.03) $~~$& 	 \scriptsize\tt 0.28 (0.02) $~$ &	 \scriptsize\tt 0.19 (0.08) $~~$& 	 \scriptsize\tt 0.41 (0.02) $~$ &	 \scriptsize\tt 0.28 (0.06) $~~$\\
      $\lambda_{S}$  & 	 \scriptsize\tt 0.147 (0.003) $~$ &	 \scriptsize\tt 0.43 (0.03) $~~$& 	 \scriptsize\tt 0.49 (0.01) $~$ &	 \scriptsize\tt 0.50 (0.03) $~~$& 	 \scriptsize\tt 0.35 (0.01) $~$ &	 \scriptsize\tt 0.45 (0.05) $~~$\\
      \bottomrule
\end{tabular}
\vskip 0.2em
\caption{Results of the benchmark. SL is the Standard Learner. AULC is the Area Under the Learning Curve of the error rate. Fraction indicates the average fraction of non-monotone decisions during a single run. Standard deviation shown in (braces). Best monotonicity result is \underline{underlined}.}
\label{table_bench}
\end{table}

\subsection{Second Experiment: Benchmark on Peaking, Dipping, MNIST}

Interestingly, for the peaking and MNIST any non-monotonicity in the expected learning curve completely disappears for $\lambda_S$ that tunes the regularization parameter. However, for the dipping dataset this is not the case --- thus regularization may not always avoid non-monotone behaviour. Furthermore, the fraction of non-monotone decisions per run is largest for this learner. It is strange that for the MNIST dataset this learner starts lagging behind other learners for large sample sizes.

For the dipping dataset $\text{M}_{\text{CV}}$ has a large advantage in terms of AULC. We hypothize that this is largely due to tie breaking and small training set sizes due to the 5-folds. Surprisingly on the peaking dataset it seems to learn quite slowly. The expected learning curves of $\text{MT}_{\text{HT}}$ look better than that of  $\text{MT}_{\text{SIMPLE}}$, however, in terms of AULC the difference is quite small. 

Again the fraction of non-monotone decisions for $\text{MT}_{\text{HT}}$ per run is very small as guaranteed. However, it is interesting to note that this does not always translate to monotonicity in the expected learning curve. For example, for peaking and dipping the expected curve doesn't seem entirely monotone. But $\text{MT}_{\text{CV}}$, which makes many non-monotone decisions per run, still seems to have a monotone expected learning curve. This really does seem to indicate that monotonicity in individual curves and monotonicity in the expected curve are not necessarily related goals. This raises the question: under what conditions do we have monotonicity in the expected learning curve?

\subsection{General Remarks}

That the fraction of non-monotone decisions of $\text{MT}_{\text{HT}}$ is so much smaller than $\alpha$ may indicate the hypothesis test is too pessimistic. \citet{Fagerland2013} indicate that the asymptotic McNemar test may have more power. For this test the guarantee $P(p \leq \alpha | H_0) \leq \alpha$ can be violated, but in light of the monotonicity results we have obtained this may not be a problem in practice. The added power could further improve the AULC.

We would like to argue that possible inconsistency of $\text{MT}_{\text{HT}}$ is not so problematic. If one knows the desired error rate, this can be used to estimate a minimum $N_v$ that ensures the hypothesis test will not get stuck before reaching that error rate. Another way to get around this issue is to make the size $N_v$ dependent on $i$: if $N_v$ is monotonically increasing this directly leads to consistency of $\text{MT}_{\text{HT}}$. It would be ideal if somehow $N_v$ could be automatically tuned to trade-off sample size requirements, consistency and monotonicity. For future work we also intend to investigate how to combine $\text{MT}_{\text{HT}}$ and $\text{MT}_{\text{CV}}$, since for CV $N_v$ automatically grows and thus also directly implies consistency.

We suspect that the peak in the feature curves that \citet{Belkin2019} observe, is  due to the same peaking phenomena as seen by \citet{Duin1995}. We wonder if optimal tuning of the regularization parameter therefore eliminates the double-descent curve, as \citet{Belkin2019} calls this behaviour, as in our setup? 

\citet{Devroye1996} conjectured that it would be impossible to construct a learner that is monotone as judged by the expected learning curve that is also consistent. While our work does not disprove this conjecture, as we look at monotonicity of individual curves, some of us suspect this is a first step in that direction. First, however, we require a better understanding of the relation between monotonicity of individual curves and of the expected learning curve. 

\section{Conclusion}

We have introduced three algorithms to make learners more monotone. We proved under which conditions the algorithms are consistent and we have shown for $\text{MT}_{\text{HT}}$ that the learning curve is monotone with high probability. If one cares only about monotonicity of the expected learning curve, $\text{MT}_{\text{SIMPLE}}$ with very large $N_v$ or $\text{MT}_{\text{CV}}$ may prove sufficient as shown by our experiments. However, they come without any theoretical guarantees. If $N_v$ is small, or one desires monotonicity of individual learning curves (as practically most relevant), $\text{MT}_{\text{HT}}$ is the right choice. Our algorithms are a first step towards developing learners that, given more data, will improve their performance in expectation.

\bibliography{referencesfixed}

\begin{thebibliography}{18}
\providecommand{\natexlab}[1]{#1}
\providecommand{\url}[1]{\texttt{#1}}
\expandafter\ifx\csname urlstyle\endcsname\relax
  \providecommand{\doi}[1]{doi: #1}\else
  \providecommand{\doi}{doi: \begingroup \urlstyle{rm}\Url}\fi

\bibitem[Shalev-Shwartz and Ben-David(2014)]{shalev2014understanding}
Shai Shalev-Shwartz and Shai Ben-David.
\newblock \emph{{Understanding machine learning: From theory to algorithms}}.
\newblock Cambridge university press, 2014.

\bibitem[Duin(1995)]{Duin1995}
R.P.W. Duin.
\newblock {Small sample size generalization}.
\newblock \emph{9th SCIA}, \penalty0 (May):\penalty0 1--8, 1995.

\bibitem[Opper and Kinzel(1996)]{Opper1996}
Manfred Opper and Wolfgang Kinzel.
\newblock {Statistical mechanics of generalization}.
\newblock In \emph{Models of neural networks III}, pages 151--209. Springer,
  1996.

\bibitem[Loog and Duin(2012)]{Loog2012}
Marco Loog and R.P.W. Duin.
\newblock {The dipping phenomenon}.
\newblock In \emph{S+SSPR}, pages 310--317, Hiroshima, Japan, 2012.

\bibitem[Belkin et~al.(2019)Belkin, Hsu, Ma, and Mandal]{Belkin2019}
Mikhail Belkin, Daniel Hsu, Siyuan Ma, and Soumik Mandal.
\newblock {Reconciling modern machine-learning practice and the classical
  bias–variance trade-off}.
\newblock \emph{PNAS}, 116\penalty0 (32):\penalty0 15849--15854, 2019.

\bibitem[Spigler et~al.(2018)Spigler, Geiger, D'Ascoli, Sagun, Biroli, and
  Wyart]{Spigler2018}
Stefano Spigler, Mario Geiger, Stéphane D'Ascoli, Levent Sagun, Giulio Biroli,
  and Matthieu Wyart.
\newblock {A jamming transition from under- to over-parametrization affects
  loss landscape and generalization}.
\newblock \emph{arXiv preprint arXiv:1810.09665}, 2018.

\bibitem[Viering et~al.(2019)Viering, Mey, and Loog]{Viering2019}
Tom Viering, Alexander Mey, and Marco Loog.
\newblock Open problem: Monotonicity of learning.
\newblock In \emph{COLT}, pages 3198--3201, 2019.

\bibitem[Loog et~al.(2019)Loog, Viering, and Mey]{Loog2019}
Marco Loog, Tom Viering, and Alexander Mey.
\newblock {Minimizers of the Empirical Risk and Risk Monotonicity}.
\newblock In \emph{NeuRIPS 2019}, 2019.

\bibitem[LeCun et~al.(1998)LeCun, Bottou, Bengio, and Haffner]{Lecun1998}
Yann LeCun, Léon Bottou, Yoshua Bengio, and Patrick Haffner.
\newblock {Gradient-based learning applied to document recognition}.
\newblock \emph{Proceedings of the IEEE}, 86\penalty0 (11):\penalty0
  2278--2324, 1998.

\bibitem[O’Neill et~al.(2017)O’Neill, Delany, and MacNamee]{o2017model}
Jack O’Neill, Sarah~Jane Delany, and Brian MacNamee.
\newblock Model-free and model-based active learning for regression.
\newblock In \emph{Advances in Computational Intelligence Systems}, pages
  375--386. Springer, 2017.

\bibitem[Huijser and van Gemert(2017)]{huijser2017active}
Miriam Huijser and Jan~C van Gemert.
\newblock Active decision boundary annotation with deep generative models.
\newblock In \emph{Proceedings of the IEEE International Conference on Computer
  Vision}, pages 5286--5295, 2017.

\bibitem[Settles and Craven(2008)]{settles2008analysis}
Burr Settles and Mark Craven.
\newblock An analysis of active learning strategies for sequence labeling
  tasks.
\newblock In \emph{Proceedings of the conference on empirical methods in
  natural language processing}, pages 1070--1079. Association for Computational
  Linguistics, 2008.

\bibitem[Japkowicz and Shah(2011)]{japkowicz2011evaluating}
Nathalie Japkowicz and Mohak Shah.
\newblock \emph{{Evaluating learning algorithms: a classification
  perspective}}.
\newblock Cambridge University Press, 2011.

\bibitem[Raschka(2018)]{raschka2018model}
Sebastian Raschka.
\newblock {Model evaluation, model selection, and algorithm selection in
  machine learning}.
\newblock \emph{arXiv preprint arXiv:1811.12808}, 2018.

\bibitem[Hastie et~al.(2005)Hastie, Tibshirani, Friedman, and
  Franklin]{hastie2005elements}
Trevor Hastie, Robert Tibshirani, Jerome Friedman, and James Franklin.
\newblock \emph{{The elements of statistical learning}}.
\newblock Springer, 2005.

\bibitem[Rifkin et~al.(2003)Rifkin, Yeo, and Poggio]{rifkin2003regularized}
Ryan Rifkin, Gene Yeo, and Tomaso Poggio.
\newblock {Regularized least-squares classification}.
\newblock \emph{Nato Science Series Sub Series III Computer and Systems
  Sciences}, 190:\penalty0 131--154, 2003.

\bibitem[Fagerland et~al.(2013)Fagerland, Lydersen, and Laake]{Fagerland2013}
Morten~W Fagerland, Stian Lydersen, and Petter Laake.
\newblock {The McNemar test for binary matched-pairs data: mid-p and asymptotic
  are better than exact conditional.}
\newblock \emph{BMC medical research methodology}, 13:\penalty0 91, 7 2013.

\bibitem[Devroye et~al.(1996)Devroye, Gy{\"{o}}rfi, and Lugosi]{Devroye1996}
Luc Devroye, László Gy{\"{o}}rfi, and Gábor Lugosi.
\newblock \emph{{A Probabilistic Theory of Pattern Recognition}}, volume~31.
\newblock Springer, New York, NY, USA, 1996.

\end{thebibliography}

\end{document}